%% file: cvpr_oi.tex
\documentclass[10pt,twocolumn,letterpaper]{article}
\usepackage[pagenumbers]{cvpr}
\makeatletter
\patchcmd{\@maketitle}{\vspace*{24pt}}{\vspace*{14pt}}{}{}
\makeatother
\input{preamble}
\definecolor{cvprblue}{rgb}{0.21,0.49,0.74}
\usepackage[pagebackref,breaklinks,colorlinks,allcolors=cvprblue]{hyperref}
\usepackage{orcidlink}
\usepackage[capitalize]{cleveref}
\hypersetup{pdfauthor={Zijie Xin, Tianyi Wang, Daoze Zhang, Jie Yang, Changli Tang, Qijie Wei, Fengyun Rao, Xirong Li},pdftitle={O-IN: Can Omni-modal LLMs Understand Spoken Intentions?}}
\def\confName{CVPR}
\def\confYear{2027}

\newcommand{\specialcell}[2][c]{%
  \begin{tabular}[#1]{@{}c@{}}#2\end{tabular}}

\definecolor{cGrey}{HTML}{F3F7F2}
\definecolor{vcolor}{HTML}{4A90D9}
\definecolor{acolor}{HTML}{D94A4A}

\def\varinner#1,#2\relax{\,{\scriptsize(\textcolor{vcolor}{#1}{\text{-}}\textcolor{acolor}{#2})}}

\def\varappinner#1,#2\relax{\textcolor{vcolor}{#1}{\text{-}}\textcolor{acolor}{#2}}
\crefname{table}{Tab.}{Tables}
\crefname{figure}{Fig.}{Figures}
\crefname{section}{Sec.}{Sections}
\crefname{equation}{Eq.}{Equations}
\crefname{appendix}{Appendix}{Appendices}

\newcommand{\envelope}{\ding{41}}
\newcommand{\pen}{\ding{45}}
\newcommand{\cmark}{\textcolor{green!60!black}{\ding{51}}}
\newcommand{\xmark}{\textcolor{red}{\ding{55}}}

\newcommand{\ourmethod}{{\fontfamily{lmtt}O-IN}\xspace}
\newcommand{\ourbench}{{\fontfamily{lmtt}\ourmethod-Bench}\xspace}
\newcommand{\ourengine}{{\fontfamily{lmtt}\ourmethod-Engine}\xspace}

\title{
\raisebox{-0.4em}{\includegraphics[height=1.4em]{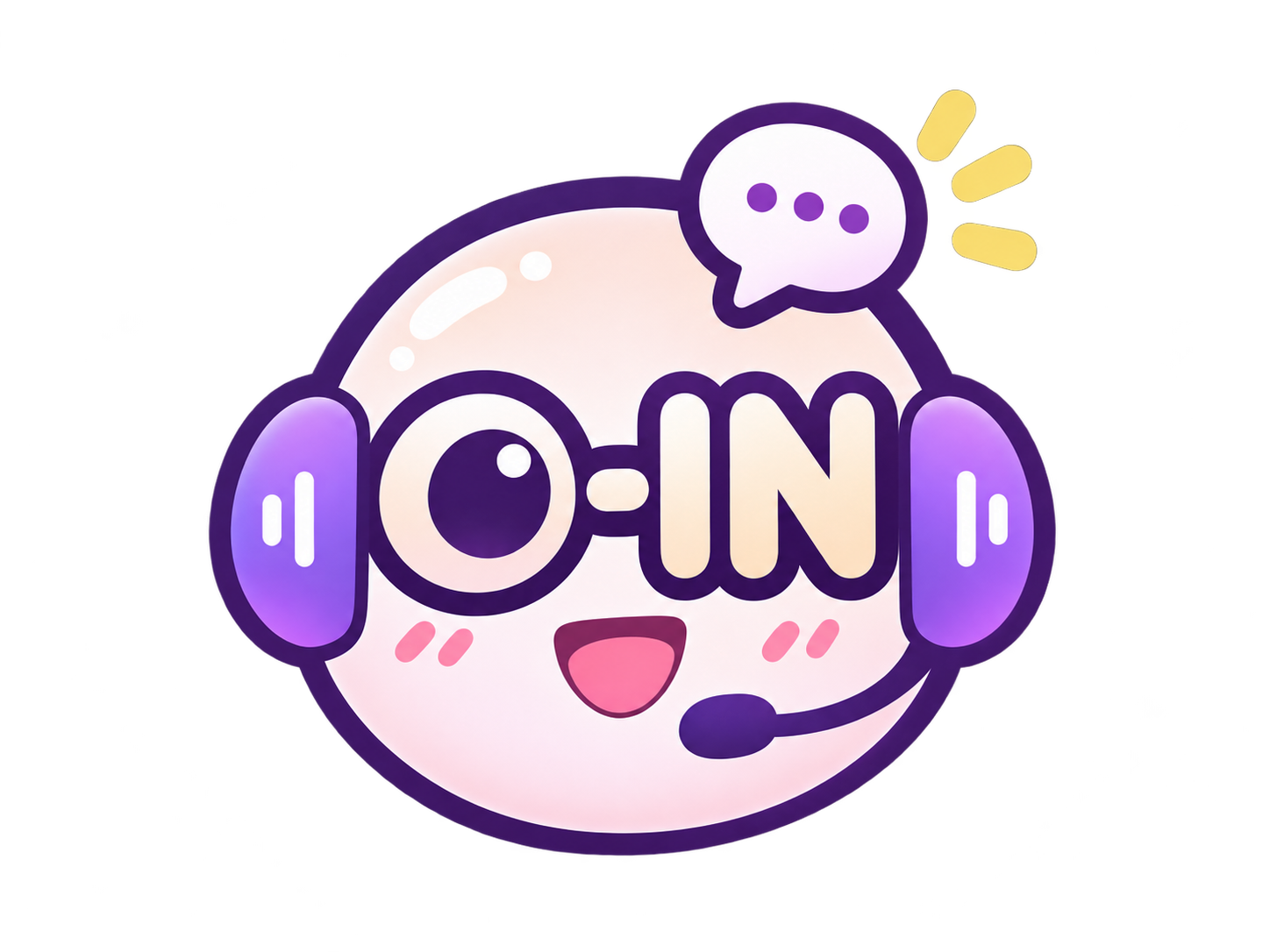}}
Can Omni-modal LLMs Understand Spoken Intentions?}
\newcommand{\authorname}[2]{\mbox{#1\textsuperscript{\textmd{#2}}}}
\author{%
\begin{tabular}{@{}c@{}}
\bfseries
\authorname{Zijie Xin}{1\text{\pen}}\orcidlink{0000-0002-9220-8735} \quad
\authorname{Tianyi Wang}{2} \quad
\authorname{Daoze Zhang}{2} \quad
\authorname{Jie Yang}{2} \quad
\authorname{Changli Tang}{2} \quad
\authorname{Qijie Wei}{1} \\[1mm]
\bfseries
\authorname{Fengyun Rao}{2} \quad
\authorname{Xirong Li}{1,\text{\envelope}}\orcidlink{0000-0002-0220-8310}
\\[1.5mm]
\normalfont
$^1$~Renmin University of China \quad $^2$~WeChat Vision, Tencent Inc. \\[2mm]
\small\textbf{Project page: }\texttt{\url{https://xxayt.github.io/O-IN-Bench}}
\end{tabular}
}

\begin{document}
\twocolumn[{
    \renewcommand\twocolumn[1][]{#1}
    \maketitle
    \vspace{-29pt}
    \input{figure/fig_teaser}
}]
\renewcommand{\thefootnote}{\pen}
\footnotetext[1]{Work performed as an intern at Tencent. (xinzijie@ruc.edu.cn)}
\renewcommand{\thefootnote}{\envelope}
\footnotetext[2]{Corresponding author. (xirong@ruc.edu.cn)}
\renewcommand{\thefootnote}{\arabic{footnote}}

\input{sec/0_abstract} \label{sec:abs}
\section{Introduction} \label{sec:intro}
\input{sec/1_intro}

\section{Related Work} \label{sec:related}

\input{sec/2_related}
\section{\ourbench: A Benchmark For Spoken Intention Understanding} \label{sec:dataset}
\input{sec/3_dataset}

\section{Evaluation}\label{sec:experiment}

\input{sec/5_experiment}

\section{Conclusion} \label{sec:conc}
\input{sec/6_conclusion}

\raggedbottom

\section*{Acknowledgments}
This research was supported by NSFC (No.62576348), BJNSF (No.L254039), Tencent WeChat Rhino-Bird Focused Research Program, and the Outstanding Innovative Talents Cultivation Funded Programs 2025 of Renmin University of China.

\section*{Release Notes}
\begin{itemize}
    \item \textbf{v1: Technical Report.} This version presents \ourbench, its construction pipeline, evaluation protocol, and experimental findings. Updated versions will follow.
\end{itemize}

{\small
\bibliographystyle{ieeenat_fullname}
\bibliography{main}
}
\end{document}

%% file: preamble.tex
\usepackage[T1]{fontenc}
\usepackage{microtype}
\usepackage{multirow,colortbl,tabularx}
\usepackage{fontawesome,pifont}
\usepackage[breakable]{tcolorbox}
\usepackage{fancyvrb,fvextra}

%% file: figure/fig_teaser.tex
\begin{center}
    \includegraphics[width=\textwidth]{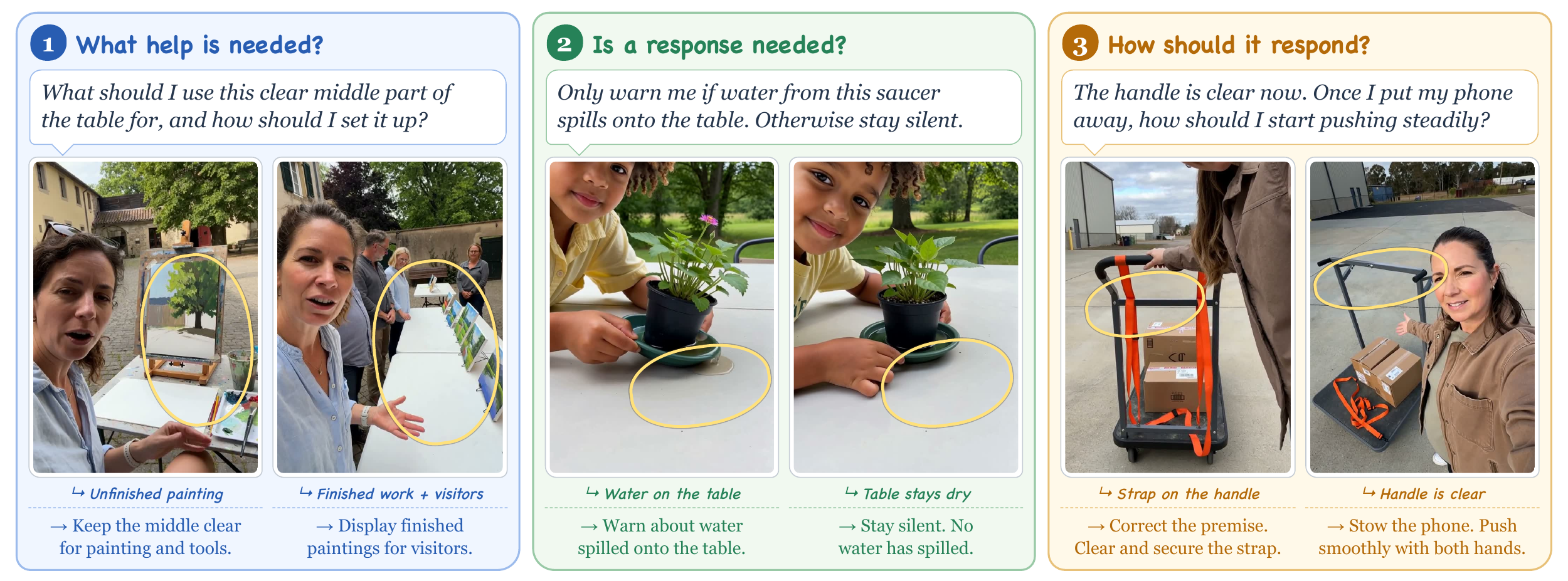}
    \captionof{figure}{\textbf{The same spoken request can require different help or response behavior.} Paired examples show how the activity context (left), whether an agreed event occurs (middle), and whether a stated premise holds (right) determine the appropriate response.}
    \label{fig:teaser}
\end{center}

%% file: sec/0_abstract.tex
\begin{abstract}
Omni-modal large language models (omni-modal LLMs) can respond to spoken requests directly through voice and video calls.
Such requests rely on the scene shared by the user and the assistant, so the same utterance may call for different help or for no response.
Most audio-visual benchmarks provide questions as separate text input, and interactive benchmarks mostly score a response as a whole.
It thus remains unclear whether a model recognizes when the user is asking for help, and whether its reply meets the user's need once the request is recognized.
We propose \ourbench to evaluate how well \underline{o}mni-modal LLMs understand spoken \underline{in}tentions.
A model watches a complete video with native speech and outputs one text response, or \texttt{[SILENCE]} when no response is needed.
\ourbench contains 1{,}000 Chinese and English samples in five categories and 16 subcategories.
Contrastive pairs combine the same request with different scene conditions to test whether the response changes with the context.
A sample-specific rubric judges language, intent, and content separately, and \ourengine verifies each rubric against the video.
Across 30 omni-modal model configurations, the best overall score is 59.6, achieved by Gemini3.7-Flash.
Gemini3.7-Flash passes 74.4\% of the intent judgments but only 42.6\% of the content judgments.
When the speech is a conversation between peers rather than a request to the assistant, even strong models often keep responding.
When a reply must satisfy several people, the best score is only 25.9.
These results indicate that recognizing a request and meeting its requirements are separate abilities, and \ourbench measures each of them.
\end{abstract}

%% file: sec/1_intro.tex
\input{table/table_benchmark}

Omni-modal large language models (omni-modal LLMs) have made remarkable progress in offline audio-visual understanding~\citep{comanici2025gemini25,xu2025qwen3omni,ye2026omnivinci}.
They can also respond to spoken requests directly through voice and video calls~\citep{hurst2024gpt4osystemcard,google2026geminiliveapi,bytedance2026seedrealtime,cui2026minicpmo45}, where the request is embedded in the audio and video without a separate text question.
Such requests rely on the scene shared with the assistant, so the same utterance may call for different help, as shown in \cref{fig:teaser}.
For example, a user who asks how to use the clear middle of a table needs a workspace while painting, but a display area when visitors are viewing the finished paintings.
A user who asks to be warned if water spills from a saucer needs a warning only when the spill occurs.
We call the help that a user seeks from the assistant a \textit{spoken intention}.
Understanding it requires deciding whether to respond and what reply meets the need.

Audio-visual understanding benchmarks~\citep{li2025omnibench,hong2026worldsense,gong2024avodyssey,zhou2025dailyomni,li2026omnivideobench,hu2026evaomni} pose questions in text to examine whether a model fully understands the audio-visual content.
A text question already specifies the request, so the model needs neither to identify who is asking for help nor to decide whether to respond.
Intent understanding benchmarks~\citep{zhang2024mintrec20,veerabadran2025wagibench} predict intent categories or user goals, which does not show whether a model can provide the needed help.

Recent interactive benchmarks place requests in speech, either as separate or synthesized audio~\citep{wang2025omnimmi,selvakumar2025multivox,sun2026omniassistbench} or as native speech from people in the scene~\citep{pourreza2026qivd,lu2026omniinteract,mazumdar2026videofdb,chen2026odubench,he2026omnivchat}.
Most of them score a response as a whole against a reference answer.
VideoFDB and ODU-Bench score separate criteria, but they evaluate conversational behavior and recovered demands rather than whether the response meets the need.
Their requests are also mostly explicit.
Unstated needs, gestural constraints, clarification, and responses under changing conditions are rarely evaluated, as discussed in \cref{sec:related}.

These gaps leave three questions open.
\textbf{Q1.} Can the model recognize when the user is asking for help?
In a real scene, the user may talk to a peer, withdraw a request, or ask the assistant to respond only after an event occurs.
The model must decide whether to respond before deciding how.
A text question removes this decision, because it always signals that an answer is expected.
\textbf{Q2.} Does recognizing the request guarantee meeting the user's need?
A response may address the correct request yet ignore a gestural constraint or contradict a visible fact.
A holistic score treats such a response like one that misses the request, so it cannot show which step fails.
\textbf{Q3.} Does the response change with the context?
The same utterance may call for different help when the activity, an agreed event, or a stated premise changes.
If each scene is evaluated under only one condition, a model that uses the context cannot be distinguished from a model that always gives the same kind of response.

To answer these questions, we propose \ourbench.
For Q1, a model watches a complete video with native speech and outputs one text response, or \texttt{[SILENCE]} when no response is needed.
Five categories organize what the model must determine, covering implicit requests, constraint integration, and clarification.
For Q2, each sample has a sample-specific rubric with three criteria on language, intent, and content, and an evaluator judges each criterion independently.
For Q3, the two samples of a contrastive pair share a scene and request but differ in one key condition that changes the appropriate response.
We construct the samples with \ourengine, which generates videos from scripts and revises annotations against each video.
\ourbench contains 1{,}000 Chinese and English samples.

We evaluate 30 omni-modal model configurations, and the best Overall score is 59.6 by Gemini3.7-Flash.
For Q1, configurations under 10B rarely recognize when no response is needed, and none exceeds 22.1\% silence accuracy.
For Q2, Gemini3.7-Flash passes 74.4\% of the intent judgments but only 42.6\% of the content judgments.
For Q3, the top five configurations mostly fall silent after a request is withdrawn.
When the speech is directed to a peer instead of the assistant, they keep the same decision in 71.1\% of such pairs.
Our contributions are threefold. \\
$\bullet$ \textbf{Problem formulation}. We formulate spoken intention understanding as deciding whether to respond and giving a reply that meets the user's need. We organize it into five categories and 16 subcategories, and design contrastive pairs. \\
$\bullet$ \textbf{Benchmark}. We build \ourbench with 1{,}000 native audio-visual samples, whose rubrics \ourengine verifies against the videos. \\
$\bullet$ \textbf{Empirical diagnosis}. Current models do not always recognize complex requests. Even after recognizing a request, a model cannot reliably provide help that meets the need.

%% file: table/table_benchmark.tex
\begin{table*}[t]
\caption{\textbf{Comparison of current spoken request benchmarks.}
}
\label{tab:benchmark_comparison}

\centering
\footnotesize
\setlength{\tabcolsep}{3pt}
\renewcommand{\arraystretch}{1.2}
\setlength{\aboverulesep}{0pt}
\setlength{\belowrulesep}{0pt}
\definecolor{reqbg}{HTML}{FFFBEA}
\definecolor{evalbg}{HTML}{FFF4EC}
\definecolor{taskbg}{HTML}{EFF7FF}
\definecolor{desbg}{HTML}{F6F2FF}
\newcommand{\grpgap}{4pt}
\newcolumntype{E}[1]{>{\columncolor{#1}[\tabcolsep][\dimexpr\tabcolsep+\grpgap\relax]}c@{\hspace{\dimexpr2\tabcolsep+\grpgap\relax}}}
\newcolumntype{H}[1]{@{}E{#1}}

\resizebox{\linewidth}{!}{%
\begin{tabular}{@{} l c @{\hspace{\dimexpr2\tabcolsep+\grpgap\relax}}
  >{\columncolor{reqbg}}c E{reqbg}
  >{\columncolor{evalbg}}c E{evalbg}
  *{4}{>{\columncolor{taskbg}}c} E{taskbg}
  >{\columncolor{desbg}}c}
\toprule

\multirow{3}{*}{\textbf{Benchmark}}
& \multirow{3}{*}{\specialcell{\textbf{Video}\\\textbf{Input}}}
& \multicolumn{2}{H{reqbg}}{\textbf{Request Format}}
& \multicolumn{2}{H{evalbg}}{\textbf{Main Evaluation}}
& \multicolumn{5}{H{taskbg}}{\textbf{Task Coverage}}
& \multicolumn{1}{@{}>{\columncolor{desbg}}c}{\textbf{Design}} \\
\cmidrule(r){3-4} \cmidrule(r){5-6} \cmidrule(r){7-11} \cmidrule{12-12}

& & \specialcell{\textit{In-video}\\\textit{Spoken}}
    & \specialcell{\textit{Native}\\\textit{Spoken}}
& \specialcell{\textit{Criterion}\\\textit{Scoring}}
    & \specialcell{\textit{User Need}\\\textit{Fulfillment}}
& \specialcell{\textit{Implicit}\\\textit{Requests}}
    & \specialcell{\textit{Constraint}\\\textit{Integration}}
    & \specialcell{\textit{Info.}\\\textit{Filtering}}
    & \specialcell{\textit{Appropriate}\\\textit{Silence}}
    & \specialcell{\textit{Asking for}\\\textit{Clarification}}
& \specialcell{\textit{Contrastive}\\\textit{Pairs}} \\
\midrule




Full-Duplex-Bench-v2~{\scriptsize\citep{lin2026fullduplexbenchv2}} & \xmark &
\xmark & \xmark &
\cmark & \cmark &
\xmark & \xmark & \xmark & \xmark & \xmark &
\xmark \\

OmniMMI~{\scriptsize\citep{wang2025omnimmi}} & \cmark &
\xmark & \xmark &
\xmark & \cmark &
\xmark & \xmark & \cmark & \cmark & \xmark &
\xmark \\

MultiVox~{\scriptsize\citep{selvakumar2025multivox}} & \cmark &
\xmark & \xmark &
\xmark & \cmark &
\xmark & \cmark & \xmark & \xmark & \xmark &
\cmark \\

OmniAssistBench~{\scriptsize\citep{sun2026omniassistbench}} & \cmark &
\cmark & \xmark &
\xmark & \cmark &
\xmark & \cmark & \cmark & \cmark & \xmark &
\xmark \\

QIVD~{\scriptsize\citep{pourreza2026qivd}} & \cmark &
\cmark & \cmark &
\xmark & \cmark &
\xmark & \xmark & \xmark & \xmark & \xmark &
\xmark \\

OmniInteract~{\scriptsize\citep{lu2026omniinteract}} & \cmark &
\cmark & \cmark &
\xmark & \cmark &
\xmark & \xmark & \xmark & \cmark & \xmark &
\xmark \\

VideoFDB~{\scriptsize\citep{mazumdar2026videofdb}} & \cmark &
\cmark & \cmark &
\cmark & \xmark &
\xmark & \xmark & \cmark & \cmark & \xmark &
\xmark \\

ODU-Bench~{\scriptsize\citep{chen2026odubench}} & \cmark &
\cmark & \cmark &
\cmark & \xmark &
\cmark & \xmark & \cmark & \cmark & \xmark &
\xmark \\

OmniVChat~{\scriptsize\citep{he2026omnivchat}} & \cmark &
\cmark & \cmark &
\cmark & \cmark &
\xmark & \xmark & \cmark & \cmark & \xmark &
\xmark \\[2pt]

\midrule
\textbf{\ourbench} (\textit{ours}) & \cmark &
\cmark & \cmark &
\cmark & \cmark &
\cmark & \cmark & \cmark & \cmark & \cmark &
\cmark \\

\bottomrule
\end{tabular}
}
\end{table*}

%% file: sec/2_related.tex
Table~\ref{tab:benchmark_comparison} compares benchmarks that evaluate responses to spoken requests.
These benchmarks first differ in how a request reaches the model.
OmniMMI~\citep{wang2025omnimmi} and MultiVox~\citep{selvakumar2025multivox} provide an audio query separately from the video.
OmniAssistBench~\citep{sun2026omniassistbench} adds synthesized speech to the video track, so its requests are spoken in the video.
QIVD~\citep{pourreza2026qivd}, OmniInteract~\citep{lu2026omniinteract}, VideoFDB~\citep{mazumdar2026videofdb}, ODU-Bench~\citep{chen2026odubench}, and OmniVChat~\citep{he2026omnivchat} use native spoken requests, which subjects in the scene say as part of the scene.
With native speech, the model must locate the request among other speech and background sounds.
Full-Duplex-Bench-v2~\citep{lin2026fullduplexbenchv2} studies multi-turn dialogue in speech only, without video input.

The benchmarks also differ in their main evaluation.
QIVD, OmniMMI, MultiVox, OmniAssistBench, and OmniInteract evaluate user need fulfillment, namely whether the response meets the user's need for assistance.
They score the response as a whole against a reference answer.
A holistic score does not show whether a failure comes from identifying the request or from the response content.
Criterion scoring judges a response on separate criteria instead.
VideoFDB and ODU-Bench use criterion scoring.
However, VideoFDB scores conversational behavior, and ODU-Bench scores the recovered demand instead of a response, so neither evaluates user need fulfillment.
OmniVChat and Full-Duplex-Bench-v2 combine criterion scoring with user need fulfillment.
\ourbench also combines the two and judges language, intent, and content separately.
The intent judgment checks whether the response addresses the correct request, and the content judgment checks whether the response satisfies the constraints in the scene.

Task Coverage lists five challenges that decide an appropriate response.
Information filtering excludes irrelevant speech, sounds, or actions, as when OmniMMI filters noise utterances.
Appropriate silence requires the model to stay silent when no response is needed, as with the keep-quiet output of OmniAssistBench.
These two challenges are the most widely covered.
Constraint integration combines the spoken request with constraints that are conveyed visually or by nonverbal sound, and only MultiVox and OmniAssistBench evaluate it.
Implicit requests require inferring help that the user does not state directly, and only ODU-Bench evaluates them.
No benchmark evaluates asking for clarification, where the assistant asks the user for a missing goal or missing information.
Contrastive pairs isolate the effect of a single condition on the response.
MultiVox builds Confounder Pairs that keep the text and visual input fixed and vary only vocal attributes.
\ourbench covers all five challenges with native spoken requests.
Its contrastive pairs share a scenario and vary one key condition, so the expected response changes with that condition.


%% file: sec/3_dataset.tex
\input{figure/fig_benchmark_overview}
\subsection{Task Definition}
\label{sec:task_definition}

\ourbench evaluates whether an omni-modal assistant understands spoken intentions in video-call scenarios.
A request may be stated directly or embedded in the surrounding conversation.
The model is given a video $v$ with native speech and a shared instruction. It combines utterances, actions, and the scene to identify the help that the user needs. It then outputs one text response $y$, or \texttt{[SILENCE]} when no response is required.
Each sample is a tuple $(v, T, q, c, R)$. Apart from $v$, its elements are annotations for construction and scoring, and they are not given to the model. The transcript $T$ records all speech in $v$. The target utterance $q$ is the utterance in $T$ that asks the assistant for help. For some samples that require silence, no utterance addresses the assistant, and $q$ is empty. The scene context $c$ describes the verified scene facts and events relevant to the request. The sample-specific rubric $R$ specifies the criteria that $y$ must meet (\cref{sec:eval_protocol}). Each sample also records its language, response requirement, and a reference response.

\cref{fig:sample_example} shows a Background Information Filtering (BIF) sample. A wet watercolor painting sits on a tilted board, and paint flows down its surface. Off-screen, the instructor says, ``Watch how the paint flows. Keep the board tilted.'' A peer adds, ``Lin, it's raining outside.'' The user then faces the camera and asks, ``What should I do next with this wet painting?'' The transcript $T$ contains all three utterances, and $q$ is the user's question. The instructor's line states the goal of the current activity and constrains the advice. The peer's weather remark is not a request and does not affect how the painting should be handled.
An appropriate response keeps the board tilted so that the user can continue observing the paint flow. This requirement depends on the spoken request, the relevant background speech, and the visible state of the board.

Each sample involves two judgments. The response decision determines whether the assistant should respond. The model must separate requests to the assistant from conversations among other people, and check whether a request has been withdrawn, revised, or conditioned on an event that has not occurred. When a response is needed, the model selects the reply form from the user's goal and the available evidence. The reply forms are direct assistance, clarification, qualified conclusion, and premise correction. The painting sample calls for direct assistance.


\subsection{Taxonomy}
\label{sec:intro_benchmark}

We divide spoken intention understanding into five categories by the response element that the model must determine.
\textbf{Interaction Management} decides whom to respond to, when to respond, and which background speech matters.
\textbf{Situated Request Interpretation} grounds a request in the current scene or a prior event to find the objects, states, and relations it involves.
\textbf{Implicit Intention Reasoning} infers unstated needs and attitudes from indirect statements of trouble, the ongoing activity, vocal delivery, and facial expressions.
\textbf{Request Integration} combines a revised request with retained requirements, gestural constraints, and the needs of other people.
\textbf{Response Deliberation} selects the reply form by whether the goal is clear, the evidence is sufficient, and the premise holds.
Within each category, subcategories differ in the cue or condition they test, giving 16 subcategories in total (\cref{fig:taxonomy_pie}). For example, Response Deliberation separates goal ambiguity, insufficient evidence, and false premises.
We predefine each subcategory by the cue or condition it tests and by the cases that do not qualify. For example, Request Disambiguation requires a truly unclear goal, and a request that merely allows several methods does not qualify. Full definitions and examples are in the supplementary material.

We build contrastive pairs to test whether a model adjusts its response to the context. A contrastive pair keeps the core request in a shared scenario and varies one key condition, so that the appropriate response changes, as shown in \cref{fig:teaser}.

\input{figure/fig_data_statistics}

\subsection{\ourengine}
\label{sec:data_construction}

We develop \ourengine to generate videos of people seeking assistance in everyday scenes, each annotated with a reference response and a sample-specific rubric.
As shown in \cref{fig:engine}, the engine consists of a \textit{scenario designer}, a \textit{video generator}, and a \textit{data curator}.

The input is a scene slot that specifies the subcategory, condition, response requirement, language, activity setting, characters, and camera view. The predefined subcategory definition accompanies the scene slot.
The scenario designer first identifies the challenge of the scene from the subcategory definition, including likely misjudgments and the key evidence for a correct response. It then writes a script $s$ that arranges characters, initial states, actions, and dialogue, so that each piece of evidence appears as a visible or audible event. From the script, it drafts the target utterance $\hat{q}$, scene context $\hat{c}$, rubric $\hat{R}$, and a reference response. Within a scenario group, the contrastive versions share the scene setup and differ in the designated condition.
The video generator renders each script into a video $v$ with native speech, environmental sound, and character actions.

The data curator checks each video and finalizes its annotations through three roles.
The \textit{video observer} watches the video independently and records an observation $o$ with a scene caption, an event timeline, the speech transcript $T$, and media quality checks.
The \textit{evidence verifier} examines the video with $o$, $s$, and the subcategory definition. It checks whether the addressees, key conditions, and scoring evidence hold in the video.
The \textit{annotation reviser} corrects $\hat{q}$, $\hat{c}$, and $\hat{R}$ according to the verification findings and sends them back for re-review.
A sample that passes verification receives the finalized $q$, $c$, and $R$, and a sample that fails is discarded. With $v$ and $T$, the finalized $q$, $c$, and $R$ form the sample $(v, T, q, c, R)$ defined in \cref{sec:task_definition}.

\ourengine thus places the evidence required by each subcategory in the video and checks every rubric against what the video shows. Model identities, prompts, quality checks, and the scope of manual review are in the supplementary material.


As shown in~\cref{fig:data_statistics}, \ourbench contains 1{,}000 Chinese and English samples in 16 subcategories, with 27--84 samples per subcategory. The samples cover five scene domains and three camera views, namely egocentric, handheld selfie, and fixed. Multiple speakers appear in 20.2\% of the samples. The 11.3\% of samples that require silence test when the assistant should refrain from responding.


\input{table/table_omniintent_overall}

\subsection{Evaluation Protocol}
\label{sec:eval_protocol}

We evaluate whether a model identifies the help that the user needs and responds appropriately.
The model receives $v$ and outputs $y$ or \texttt{[SILENCE]} as defined in \cref{sec:task_definition}, and the finalized $q$, $c$, and $R$ are used only to score $y$.

\paragraph{Criterion-Level Judgment.}
The sample-specific rubric $R = (R_L, R_I, R_C)$ has three criteria. $R_L$ requires the response to mainly use the user's language. $R_I$ requires the response to address the correct person and request with an appropriate reply form. $R_C$ requires the response to be consistent with scene facts, satisfy the necessary constraints, and provide feasible help.
For samples that require a response, a text-based evaluator reads the sample language, $q$, and $c$, and judges $y$ against each criterion independently. The judgments are binary, $L, I, C \in \{0, 1\}$. The rubrics accept different wording and alternative feasible solutions, and the evaluator does not receive the reference response.
In the painting sample, $R_I$ requires following the classroom objective without treating the weather remark as a request. $R_C$ requires keeping the board tilted and the paint flow visible.

\paragraph{Sample-Level Scoring.}
For samples that require a response, the sample score is $r = LI(1+C)/2$. Satisfying language and intent earns half credit, and satisfying content as well earns full credit. For samples that require silence, $r=1$ for an exact \texttt{[SILENCE]} output and $r=0$ otherwise. Overall is the unweighted mean of $r$ over all samples on a 0--100 scale. We also report per-subcategory scores, the pass rate of each judgment, and response and silence diagnostics. The supplementary material details the evaluator, its instructions, and the aggregation.

Separate judgments distinguish a response that misses the request from a response that addresses the request but fails its requirements, which a holistic score cannot do.

%% file: figure/fig_benchmark_overview.tex
\begin{figure*}[t]
    \centering
    \begin{subfigure}[t]{0.48\textwidth}
        \centering
        \includegraphics[width=\linewidth]{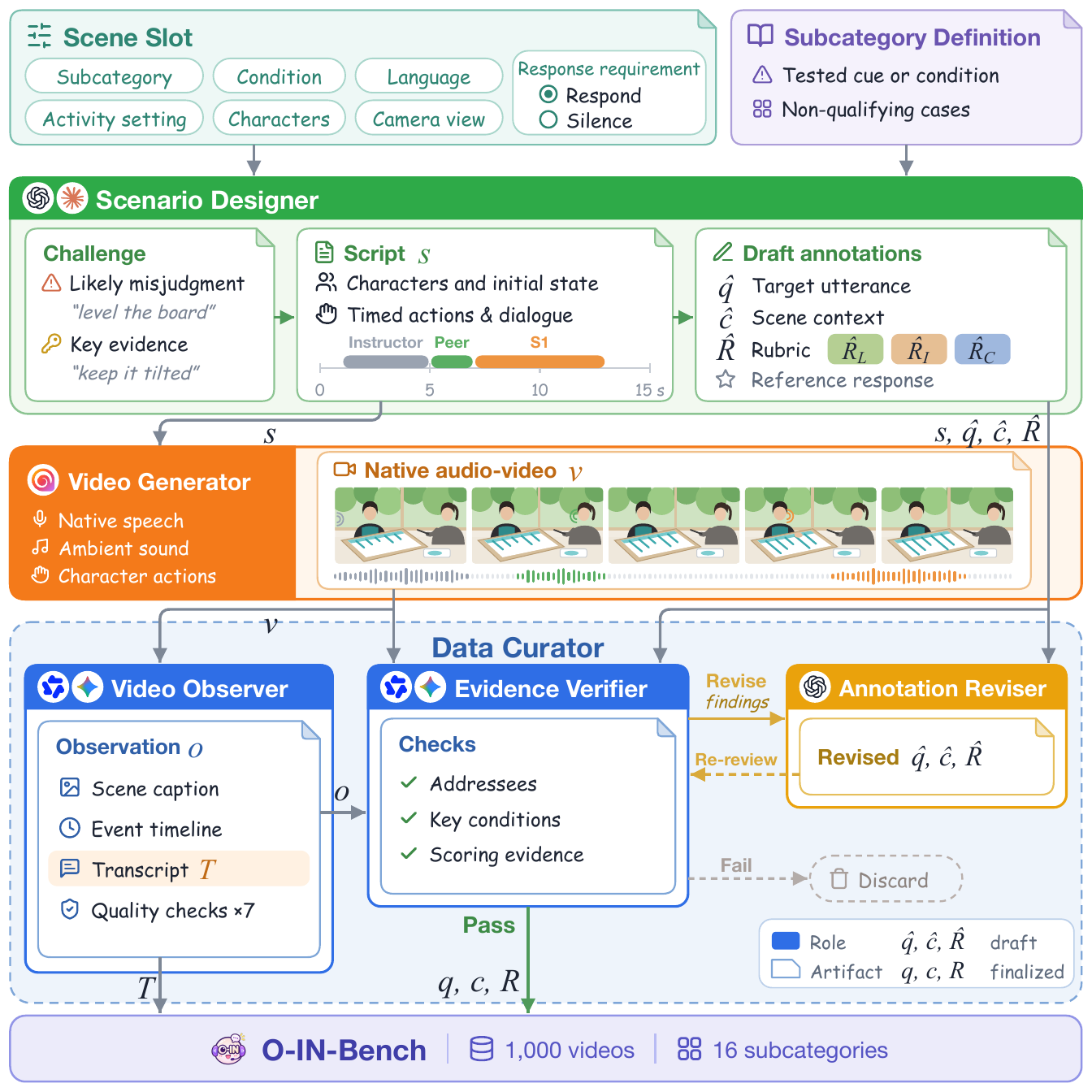}
        \caption{\ourengine.}
        \label{fig:engine}
    \end{subfigure}
    \hfill
    \begin{subfigure}[t]{0.48\textwidth}
        \centering
        \includegraphics[width=\linewidth]{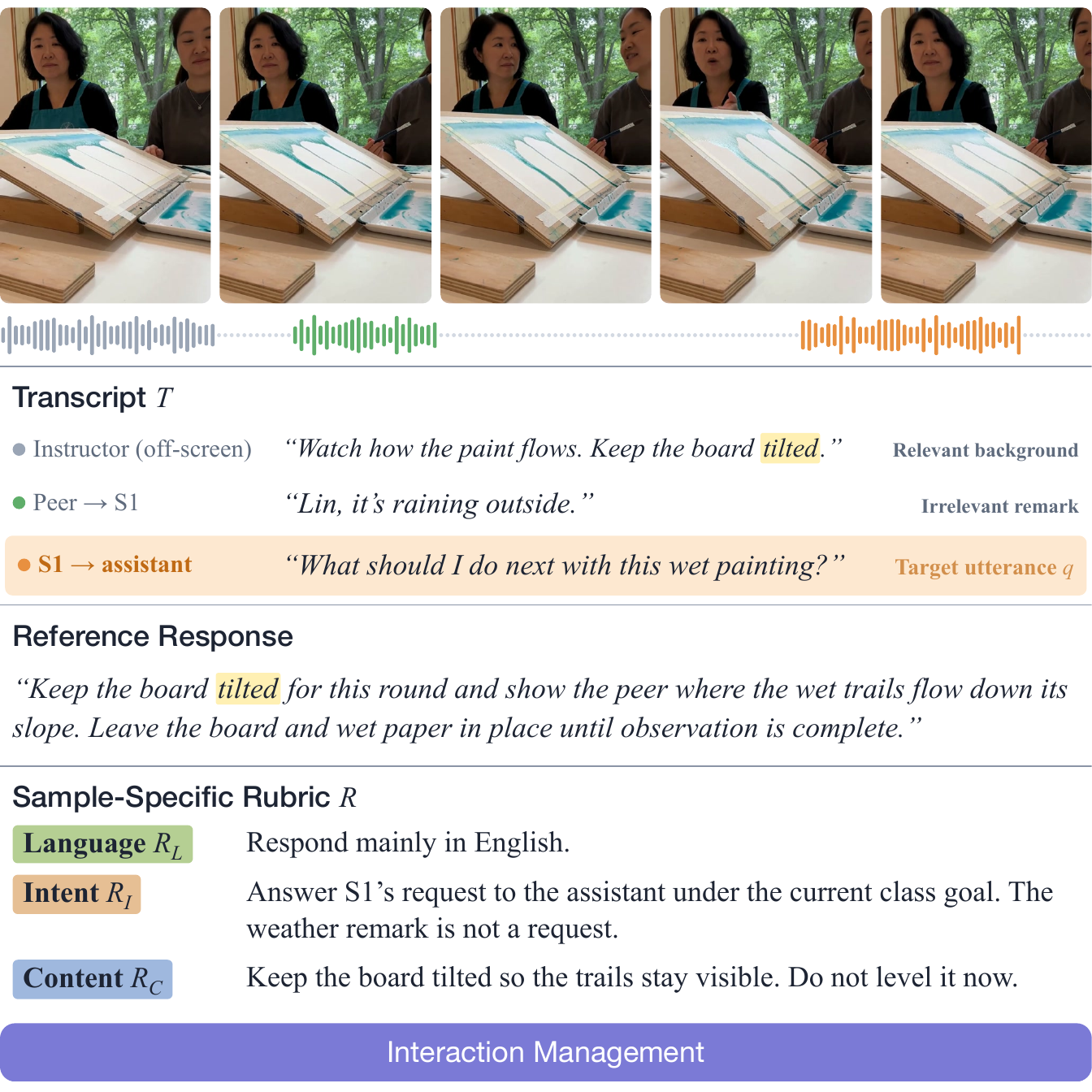}
        \caption{A sample which tests background speech filtering.}
        \label{fig:sample_example}
    \end{subfigure}
    \caption{\textbf{\ourbench overview.} (a) \ourengine writes a script $s$ and draft annotations from a scene slot and a subcategory definition, renders $s$ into a native audio-video $v$, and verifies $v$ to finalize the sample $(v, T, q, c, R)$. (b) A sample from \ourbench.}
    \label{fig:benchmark_overview}
\end{figure*}

%% file: figure/fig_data_statistics.tex
\begin{figure}[t]
    \centering
    \begin{subfigure}[t]{\linewidth}
        \centering
        \includegraphics[width=0.85\linewidth]{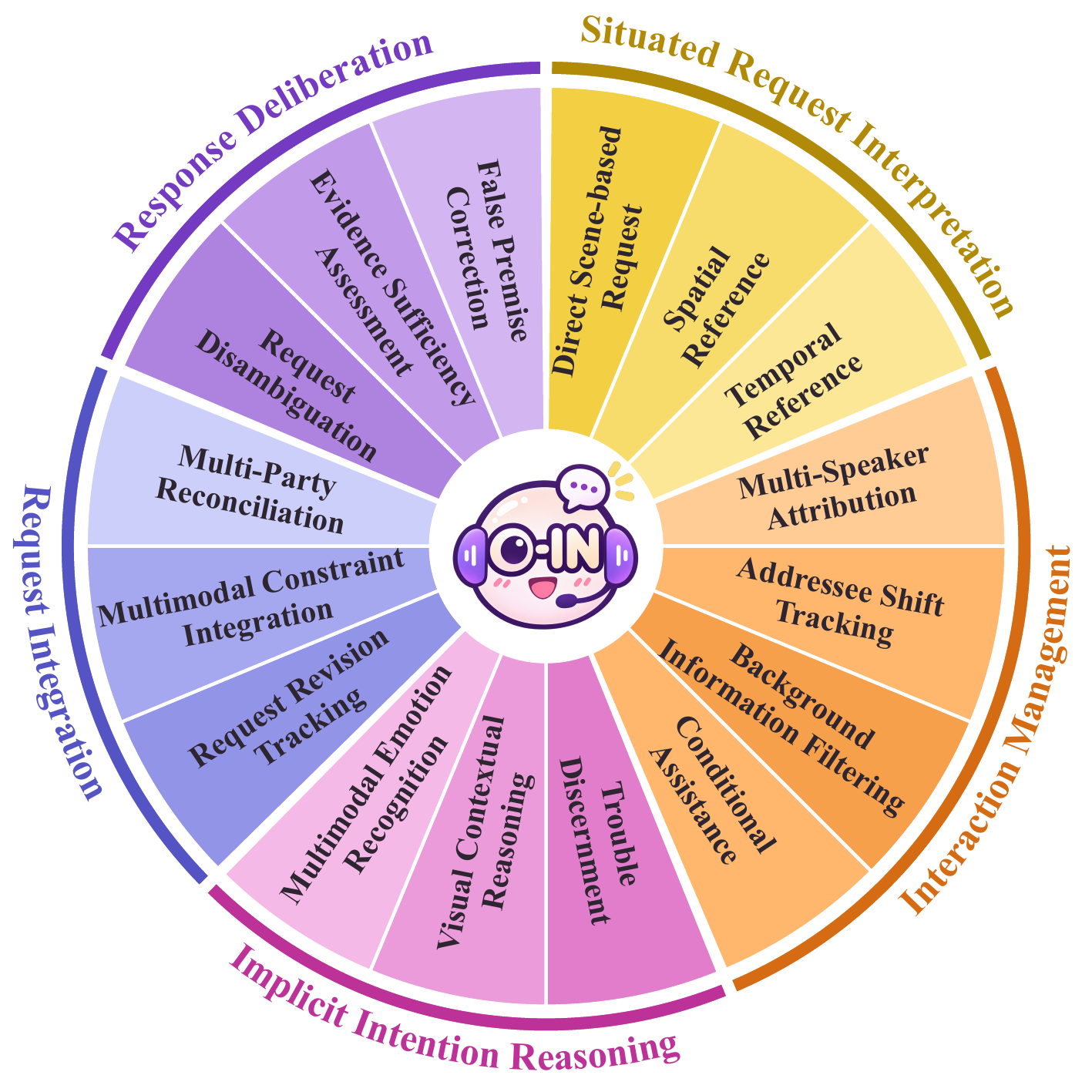}
        \caption{Task taxonomy.}
        \label{fig:taxonomy_pie}
    \end{subfigure}

    \vspace{0.5em}

    \begin{subfigure}[t]{\linewidth}
        \centering
        \includegraphics[width=\linewidth]{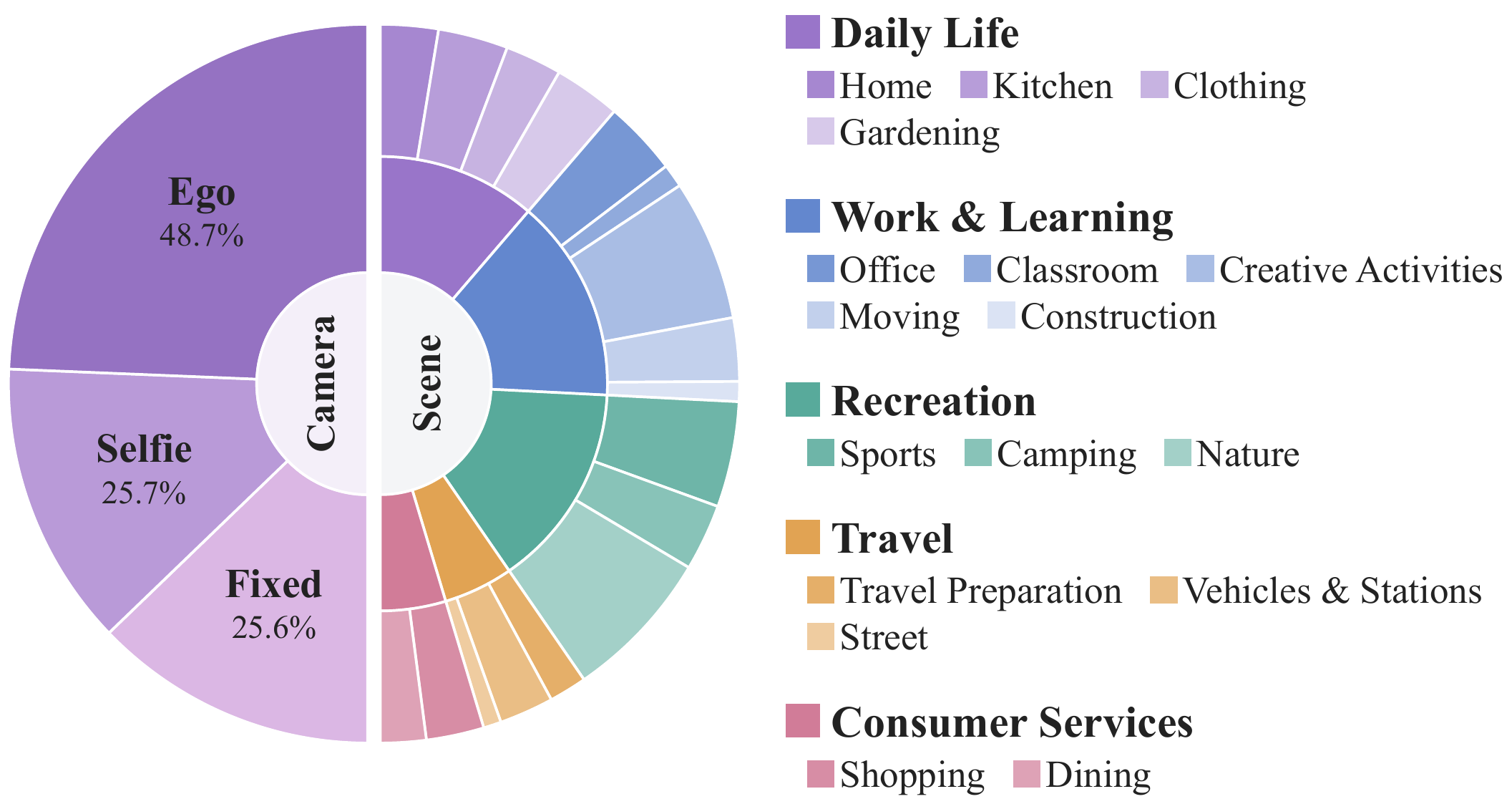}
        \caption{Scene domains and camera views.}
        \label{fig:stat_scene}
        \label{fig:stat_camera}
    \end{subfigure}
    \caption{\textbf{Task taxonomy and data statistics.} (a) Five categories and 16 subcategories. (b) Scene domains and camera views.}
    \label{fig:data_statistics}
\end{figure}

%% file: table/table_omniintent_overall.tex
\begin{table*}[t]
\centering
\newcommand{\tablemodellogo}[1]{%
  \raisebox{0.35em}[0pt][0pt]{\makebox[1.15em][c]{\raisebox{-0.5\height}{#1}}}\hspace{0.2em}%
}
\newcommand{\modelicon}[1]{\tablemodellogo{\includegraphics[width=1.1em,height=0.95em,keepaspectratio]{figure/icon/#1.png}}}
\caption{\textbf{Overall performance on \ourbench.} Scores are sample averages.
\textbf{Bold} and \underline{underline} denote the best and second-best per column within each group.
Proprietary and \textsuperscript{\textdagger}-marked models run in default thinking mode.}
\label{tab:omniintent_overall}
\footnotesize
\setlength{\tabcolsep}{3pt}
\renewcommand{\arraystretch}{1.1}

\resizebox{\linewidth}{!}{%
\begin{tabular}{@{}l *{3}{>{\centering\arraybackslash}p{2.2em}} c *{4}{>{\centering\arraybackslash}p{2.2em}} c *{3}{>{\centering\arraybackslash}p{2.2em}} c *{3}{>{\centering\arraybackslash}p{2.2em}} c *{3}{>{\centering\arraybackslash}p{2.2em}} r@{}}
\toprule
\multirow{2}{*}{\textbf{Model}} & 
\multicolumn{3}{c}{\specialcell{\textbf{Situated Request}\\\textbf{Interpretation}}} & & 
\multicolumn{4}{c}{\specialcell{\textbf{Interaction}\\\textbf{Management}}} & & 
\multicolumn{3}{c}{\specialcell{\textbf{Implicit Intention}\\\textbf{Reasoning}}} & & 
\multicolumn{3}{c}{\specialcell{\textbf{Request}\\\textbf{Integration}}} & & 
\multicolumn{3}{c}{\specialcell{\textbf{Response}\\\textbf{Deliberation}}} & \multirow{2}{*}{\textbf{Overall}} \\

\cmidrule{2-4}\cmidrule{6-9}\cmidrule{11-13}\cmidrule{15-17} \cmidrule{19-21}
& DSR & SR & TR & & 
MSA & AST & BIF & CA & & 
TD & VCR & MER & & 
RRT & MCI & MPR & & 
RD & ESA & FPC & \\
\midrule

\multicolumn{22}{c}{\cellcolor{blue!10}\textit{Proprietary Omni-modal LLMs}} \\

\modelicon{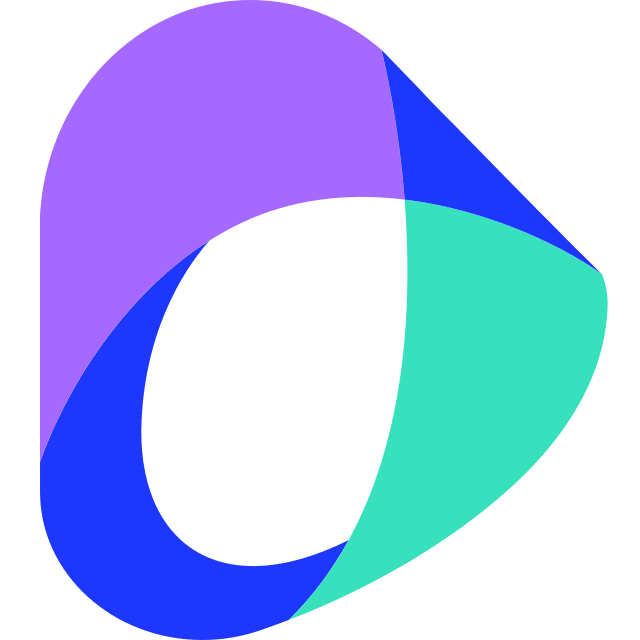}Doubao-Seed2.0-Mini-260428 & 65.5 & 55.6 & 62.5 & $ $ & 39.9 & 52.4 & 42.9 & 72.6 & $ $ & 24.4 & \underline{61.4} & \textbf{80.2} & $ $ & 51.2 & 32.6 & 9.3 & $ $ & 38.4 & 40.2 & 44.6 & 49.4 \\
\modelicon{bytedance}Doubao-Seed2.0-Lite-260428 & 60.1 & 53.3 & 58.0 & $ $ & 28.4 & 62.7 & 47.1 & 70.2 & $ $ & 55.4 & 44.6 & 51.9 & $ $ & 56.7 & 34.9 & 7.4 & $ $ & 34.8 & 34.8 & 36.1 & 47.6 \\

\modelicon{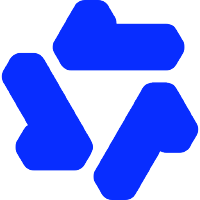}Qwen3.5-Omni-Flash & 45.3 & 42.2 & 42.0 & $ $ & 13.5 & 17.5 & 37.1 & 32.3 & $ $ & 39.9 & 43.4 & \underline{74.5} & $ $ & 39.6 & 18.6 & 3.7 & $ $ & 24.4 & 38.0 & 30.1 & 34.2 \\
\modelicon{qwen}Qwen3.8-Omni-Flash & 73.0 & 57.8 & 58.0 & $ $ & 46.6 & 78.9 & 55.7 & 79.0 & $ $ & 73.2 & 53.0 & 47.2 & $ $ & 64.6 & 43.0 & \textbf{25.9} & $ $ & 43.9 & 37.0 & 44.6 & 57.3 \\

\modelicon{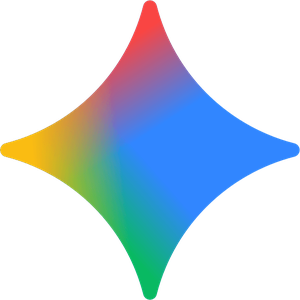}Gemini2.5-Flash-Lite & 48.0 & 46.7 & 33.0 & $ $ & 33.8 & 47.6 & 25.7 & 59.7 & $ $ & 14.9 & 38.0 & 57.5 & $ $ & 35.4 & 11.6 & 1.9 & $ $ & 36.0 & 29.3 & 35.5 & 36.3 \\
\modelicon{gemini}Gemini2.5-Flash & 57.4 & 47.8 & 47.7 & $ $ & 43.2 & 67.5 & 41.4 & 67.7 & $ $ & 32.1 & 56.0 & 69.8 & $ $ & 56.1 & 18.6 & 3.7 & $ $ & 38.4 & 35.9 & 41.0 & 47.7 \\
\modelicon{gemini}Gemini2.5-Pro & 65.5 & 43.3 & 50.0 & $ $ & 31.8 & 60.8 & 32.9 & 62.1 & $ $ & 55.4 & 31.3 & 69.8 & $ $ & 55.5 & 22.1 & 14.8 & $ $ & 42.1 & 41.3 & 37.3 & 46.7 \\
\modelicon{gemini}Gemini3-Flash-Preview & 68.9 & 58.9 & 51.1 & $ $ & 34.5 & 62.7 & 44.3 & 58.1 & $ $ & 76.2 & 56.6 & 67.0 & $ $ & 61.0 & 33.7 & 9.3 & $ $ & 43.3 & \textbf{44.6} & 41.0 & 53.3 \\
\modelicon{gemini}Gemini3-Pro-Preview & 67.6 & 60.0 & 59.1 & $ $ & 54.1 & 79.5 & 44.3 & 76.6 & $ $ & 60.7 & 54.2 & 54.7 & $ $ & 68.3 & \textbf{44.2} & 14.8 & $ $ & 47.0 & 32.6 & 44.6 & 56.7 \\  %
\modelicon{gemini}Gemini3.1-Flash-Lite & 54.7 & 44.4 & 46.6 & $ $ & 28.4 & 48.8 & 34.3 & 49.2 & $ $ & 58.9 & 50.0 & 66.0 & $ $ & 39.6 & 19.8 & 7.4 & $ $ & 36.6 & 42.4 & 42.8 & 43.9 \\
\modelicon{gemini}Gemini3.1-Pro-Preview & 67.6 & 51.1 & \underline{65.9} & $ $ & \underline{58.8} & \textbf{81.3} & 54.3 & 78.2 & $ $ & 56.5 & 50.6 & 50.9 & $ $ & 68.3 & 36.0 & 13.0 & $ $ & 53.0 & 32.6 & 46.4 & 56.9 \\
\modelicon{gemini}Gemini3.5-Flash & 72.3 & \underline{60.0} & 58.0 & $ $ & 44.6 & 75.9 & 48.6 & 81.5 & $ $ & \textbf{78.0} & 52.4 & 56.6 & $ $ & \underline{68.3} & 33.7 & 11.1 & $ $ & 52.4 & \underline{42.4} & 42.8 & 58.0 \\
\modelicon{gemini}Gemini3.6-Flash & 70.3 & 57.8 & 60.2 & $ $ & 52.7 & 77.7 & 54.3 & 79.8 & $ $ & 69.0 & 47.0 & 50.0 & $ $ & 66.5 & 30.2 & 18.5 & $ $ & 47.0 & 31.5 & 35.5 & 55.5 \\
\modelicon{gemini}Gemini3.7-Flash & \textbf{75.0} & 58.9 & \textbf{65.9} & $ $ & 44.6 & 80.1 & 55.7 & 81.5 & $ $ & 76.2 & \textbf{61.4} & 44.3 & $ $ & \textbf{70.1} & 41.9 & \underline{22.2} & $ $ & 47.6 & 37.0 & 47.0 & \textbf{59.6} \\
\modelicon{gemini}Gemini3.8-Flash & \underline{73.6} & \textbf{60.0} & 56.8 & $ $ & 47.3 & 79.5 & \underline{55.7} & \underline{81.5} & $ $ & \underline{76.8} & 52.4 & 47.2 & $ $ & 65.2 & \underline{43.0} & 16.7 & $ $ & \underline{53.0} & 39.1 & \underline{47.0} & \underline{58.8} \\
\modelicon{gemini}Gemini4-Argon & 67.6 & 46.7 & 55.7 & $ $ & \textbf{65.5} & \underline{80.1} & \textbf{55.7} & \textbf{92.7} & $ $ & 69.0 & 37.3 & 33.0 & $ $ & 64.0 & 34.9 & 14.8 & $ $ & \textbf{61.0} & 39.1 & \textbf{57.8} & 58.2 \\

\midrule

\multicolumn{22}{c}{\cellcolor{orange!10}\textbf{\textit{Open-source Omni-modal LLMs}}} \\

\modelicon{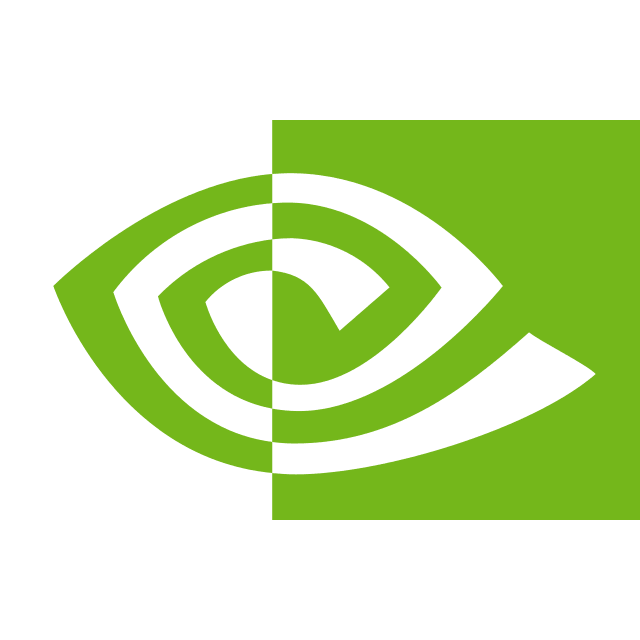}OmniVinci-9B & 0.0 & 1.1 & 0.0 & $ $ & 0.0 & 0.0 & 0.0 & 0.0 & $ $ & 0.0 & 0.0 & 0.0 & $ $ & 1.2 & 0.0 & 0.0 & $ $ & 0.6 & 0.0 & 0.0 & 0.2 \\

\modelicon{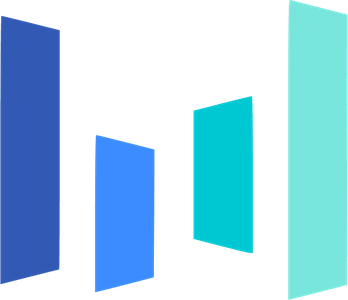}video-SALMONN 2+ 7B & 0.7 & 0.0 & 0.0 & $ $ & 0.7 & 0.0 & 5.7 & 4.0 & $ $ & 0.0 & 0.0 & 2.8 & $ $ & 1.2 & 0.0 & 0.0 & $ $ & 0.0 & 0.0 & 0.0 & 0.8 \\
\modelicon{salmonn}video-SALMONN 2+ 72B & 0.0 & 0.0 & 0.0 & $ $ & 35.1 & 38.6 & 0.0 & 58.1 & $ $ & 0.0 & 0.0 & 0.0 & $ $ & 2.4 & 0.0 & 0.0 & $ $ & 0.0 & 0.0 & 0.0 & 9.6 \\

\modelicon{nvidia}Nemotron3-Nano-Omni-30B-A3B\textsuperscript{\textdagger} & 37.8 & 25.6 & 38.6 & $ $ & 16.2 & 27.1 & 27.1 & 41.9 & $ $ & 25.0 & 25.3 & 56.6 & $ $ & 25.6 & 12.8 & 1.9 & $ $ & 25.6 & 19.6 & 16.3 & 26.9 \\

\modelicon{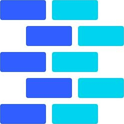}MiniCPM-o-2.6-8B & 6.1 & 3.3 & 12.5 & $ $ & 1.4 & 0.6 & 5.7 & 8.1 & $ $ & 0.6 & 6.0 & 3.8 & $ $ & 3.7 & 2.3 & 0.0 & $ $ & 3.7 & 18.5 & 1.2 & 4.4 \\
\modelicon{minicpm}MiniCPM-o-4.5-9B & 43.9 & 37.8 & 31.8 & $ $ & 12.8 & 15.1 & 31.4 & 46.0 & $ $ & 5.4 & 40.4 & 68.9 & $ $ & 34.8 & 14.0 & 3.7 & $ $ & 25.0 & 32.6 & 26.5 & 29.3 \\

\modelicon{qwen}Qwen2.5-Omni-3B & 20.9 & 17.8 & 15.9 & $ $ & 1.4 & 0.0 & 1.4 & 1.6 & $ $ & 4.8 & 19.3 & 13.2 & $ $ & 11.0 & 2.3 & 3.7 & $ $ & 14.0 & 22.8 & 13.9 & 10.5 \\
\modelicon{qwen}Qwen2.5-Omni-7B & 43.9 & 36.7 & 31.8 & $ $ & 15.5 & 17.5 & 18.6 & 3.2 & $ $ & 29.8 & 38.6 & 61.3 & $ $ & 26.2 & 19.8 & 1.9 & $ $ & 29.3 & 30.4 & 20.5 & 27.3 \\
\modelicon{qwen}Qwen3-Omni-Instruct-30B-A3B & 51.4 & 44.4 & 44.3 & $ $ & 25.0 & 24.7 & 28.6 & 14.5 & $ $ & 48.8 & 38.0 & 73.6 & $ $ & 40.2 & 17.4 & 9.3 & $ $ & 26.2 & 38.0 & 33.1 & 35.7 \\
\modelicon{qwen}Qwen3-Omni-Thinking-30B-A3B\textsuperscript{\textdagger} & 56.1 & 46.7 & 31.8 & $ $ & 28.4 & 37.3 & \textbf{54.3} & 59.7 & $ $ & 51.2 & 50.0 & 67.9 & $ $ & 52.4 & 24.4 & 13.0 & $ $ & 42.1 & \underline{39.1} & 39.8 & 44.8 \\

\modelicon{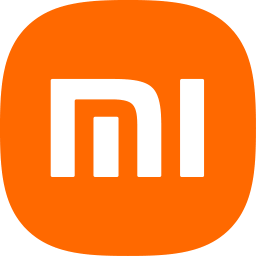}MiMo-V2.5-310B-A15B\textsuperscript{\textdagger} & 57.4 & 45.6 & 47.7 & $ $ & \textbf{40.5} & 60.8 & \underline{50.0} & 57.3 & $ $ & 58.9 & 49.4 & \textbf{83.0} & $ $ & 50.0 & 30.2 & \textbf{16.7} & $ $ & \textbf{50.6} & 37.0 & \underline{39.8} & \underline{50.2} \\
\modelicon{xiaomi}MiMo-V2.6-Flash-310B-A15B\textsuperscript{\textdagger} & \underline{58.8} & \underline{48.9} & \underline{48.9} & $ $ & \underline{36.5} & 56.6 & 44.3 & 62.1 & $ $ & 33.9 & 46.4 & \underline{77.4} & $ $ & \underline{53.0} & \underline{32.6} & 5.6 & $ $ & \underline{49.4} & \textbf{41.3} & \textbf{41.0} & 47.6 \\
\modelicon{xiaomi}MiMo-V2.6-Pro-1T-A42B\textsuperscript{\textdagger} & 58.1 & 47.8 & 46.6 & $ $ & 35.1 & \textbf{66.3} & 45.7 & \underline{72.6} & $ $ & \underline{59.5} & \underline{51.8} & 73.6 & $ $ & 51.2 & 29.1 & 7.4 & $ $ & 45.7 & 32.6 & 38.0 & 50.0 \\

\modelicon{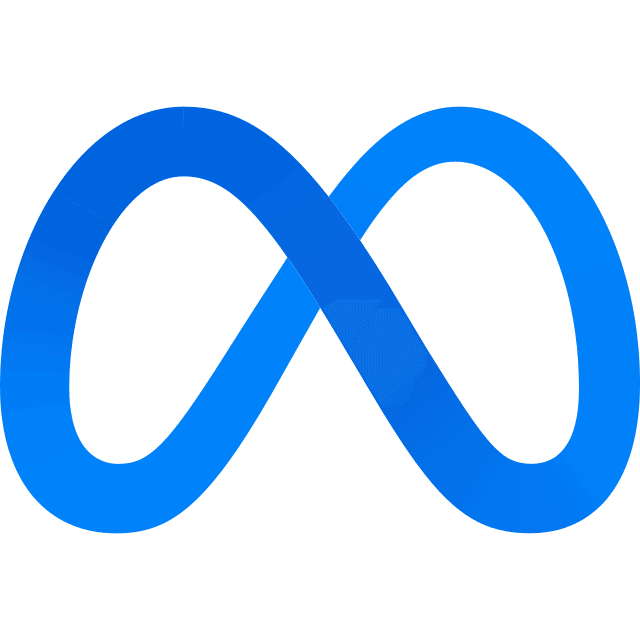}Muse-Spark-Open-816B-A42B\textsuperscript{\textdagger} & \textbf{65.5} & \textbf{56.7} & \textbf{56.8} & $ $ & 33.1 & \underline{62.7} & 45.7 & \textbf{80.6} & $ $ & \textbf{66.7} & \textbf{54.2} & 60.4 & $ $ & \textbf{59.1} & \textbf{33.7} & \underline{13.0} & $ $ & 43.9 & 33.7 & 38.0 & \textbf{52.4} \\


\bottomrule

\end{tabular}%
}
\end{table*}

%% file: sec/5_experiment.tex
\subsection{Experimental Setup}
\label{sec:eval_settings}

\textbf{Evaluated Models.}
We evaluate 30 omni-modal model configurations with native audio input, including 16 proprietary and 14 open-source configurations.
The proprietary group covers the Gemini Pro, Flash, and Flash-Lite series~\citep{comanici2025gemini25,google2025gemini3pro,google2025gemini3flash,google2026gemini31prooverview,google2026gemini31flashlite,google2026gemini35flash,google2026gemini36flash,google2026gemini37flash,google2026gemini38flash,google2026gemini4argon}, the proprietary Qwen-Omni series~\citep{team2026qwen38omni,qwen2026qwen35omni}, and Doubao-Seed2.0-Lite/Mini~\citep{bytedance2026seed20}.
The open-source group covers the Qwen-Omni series~\citep{xu2025qwen3omni,xu2025qwen25omni}, Muse-Spark-Open~\citep{meta2026musesparkopen}, MiMo-V2.5/V2.6~\citep{xiaomi2026mimov25,qiao2026mimo26}, Nemotron3-Nano-Omni~\citep{deshmukh2026nemotron3nanoomni}, OmniVinci~\citep{ye2026omnivinci}, MiniCPM-o~\citep{cui2026minicpmo45,yao2024minicpm26}, and video-SALMONN 2+~\citep{tang2025videosalmonn2}.
Tab.~\ref{tab:omniintent_overall} lists all configurations and their inference modes.
All proprietary models run in their default thinking mode.

\textbf{Implementation Details.}
We implement the evaluation with lmms-eval~\citep{zhang2025lmmseval}.
All configurations use the same 1,000 samples, with each video and its original audio as input.
Open-source models run on 8--16 NVIDIA A800 GPUs.
We serve MiMo and Nemotron with vLLM~\citep{kwon2023efficient} and Muse-Spark-Open with SGLang~\citep{zheng2024sglang}.
Proprietary models are accessed through their APIs.
We score all responses with the protocol in Sec.~\ref{sec:eval_protocol} and use GPT-6-Luna~\citep{openai2026gpt6luna} as the evaluator.
Model prompts, audiovisual preprocessing, and evaluator settings are in the supplementary material.

\subsection{Main Findings}
\label{sec:eval_overall}

\input{figure/fig_category_radar}
\input{figure/fig_3level_histogram}
\input{figure/fig_overall_language_comparison}
\input{figure/fig_silence_overall}

\input{figure/fig_modality_ablation}

Tab.~\ref{tab:omniintent_overall} reports the scores on the 16 subcategories.
Fig.~\ref{fig:category_radar} shows the five category scores of selected models, and Fig.~\ref{fig:3level_histogram} shows their language, intent, and content pass rates.

\textbf{Overall Performance.}
Gemini3.7-Flash leads the proprietary models with an Overall of 59.6.
Muse-Spark-Open leads the open-source models with 52.4, 7.2 points below Gemini3.7-Flash.
Within the Qwen3-Omni series, the thinking mode outperforms the instruct mode by 9.1 points.
The gain suggests that inferring the needed help from speech and the scene benefits from explicit reasoning.

\textbf{Performance Across Categories.}
Request Integration and Response Deliberation are the most difficult categories.
One of the two categories scores lowest for 24 of the 30 configurations (Fig.~\ref{fig:category_radar}).
Within Request Integration, 26 of the 30 configurations follow the order RRT $>$ MCI $>$ MPR.
Scores drop as a request extends from a spoken revision to nonverbal constraints and the needs of several people.
Within Response Deliberation, no configuration exceeds 44.6 on ESA, while the best DSR score is 75.0.
Judging whether the evidence supports an answer is much harder for current models than answering a direct request.

Equal Overall scores can hide different strengths.
MiMo-V2.6-Flash and Doubao-Seed2.0-Lite both score 47.6.
MiMo-V2.6-Flash is higher than Doubao-Seed2.0-Lite on all three Response Deliberation subcategories.
Doubao-Seed2.0-Lite is higher on all three Situated Request Interpretation subcategories.
Category scores thus separate interpreting a request from adjusting the reply.

\textbf{Performance Across Evaluation Dimensions.}
Content is the main bottleneck (Fig.~\ref{fig:3level_histogram}).
Gemini3.7-Flash, Qwen3.8-Omni-Flash, and Muse-Spark-Open all exceed 97\% on language and 71\% on intent, but none exceeds 43\% on content.
These models mostly address the right request in the user's language, yet their help often fails the scene facts or necessary constraints.
A higher intent pass rate does not imply a higher content pass rate.
Qwen3.8-Omni-Flash exceeds Gemini3.7-Flash by 4.6 points on intent but trails it by 10.1 points on content.
Identifying the valid request and satisfying it are thus separate abilities.

\subsection{Analysis Of Response Tendency}
\label{sec:eval_diagnostics}

\input{table/table_contrastive_pairs}

\textbf{Performance Across Languages.}
Fig.~\ref{fig:overall_language_comparison} compares Overall scores on the Chinese and English subsets.
We measure the gap with the Language Gap Index, $\mathrm{LGI}=(S_{\mathrm{en}}-S_{\mathrm{zh}})/S\times100\%$, where $S_{\mathrm{en}}$, $S_{\mathrm{zh}}$, and $S$ are the Overall scores on the English subset, the Chinese subset, and the full set.
Except for the two Flash-Lite models, all proprietary models stay within $\pm$11.2\% LGI, which may reflect broad bilingual training data.
MiMo, Qwen3-Omni, and Muse-Spark-Open show similarly small gaps.
Large gaps concentrate in smaller models such as Qwen2.5-Omni, MiniCPM-o, and Nemotron3-Nano-Omni, which often fail to reply in Chinese.
Qwen2.5-Omni-7B passes 42.2\% of language checks in Chinese and 96.8\% in English.

\textbf{Ability To Remain Silent.}
Silence accuracy rises with Overall for most models (Fig.~\ref{fig:silence_overall}).
Models that better understand spoken requests also better recognize when no response is needed.
Two exceptions stand out.
Models under 10B rarely output \texttt{[SILENCE]}, and none exceeds 22.1\% silence accuracy.
video-SALMONN 2+ 72B outputs \texttt{[SILENCE]} on most samples, reaching 83.2\% silence accuracy but only 9.6 Overall.
Its 7B version mostly describes the scene instead of answering the request.
The series is trained mainly for audiovisual captioning~\citep{tang2025videosalmonn2}, which may bias it toward scene description.

\textbf{Response Change Across Contrastive Pairs.}
Tab.~\ref{tab:contrastive_pairs} reports whether the response changes within contrastive pairs.
The top five configurations fully pass both samples in only 19.5\% of the 147 pairs, close to the 19.0\% expected if the two samples were answered independently.
Being correct under one condition thus does not make a model more likely to be correct under the other.
In the 47 pairs from MSA, AST, and CA, one sample requires a response and the other requires silence.
Always outputting \texttt{[SILENCE]} scores 50.0 on MSA, and 26 of the 30 configurations score lower.
Configurations with Overall below 36 keep the same decision in at least 78.7\% of these pairs.
The top five keep the same decision in 8.4\% of the AST pairs but 71.1\% of the MSA pairs.
They fall silent once a request is withdrawn, yet still respond to peer conversations.
With nearly equal Overall scores, Gemini4-Argon keeps the same MSA decision in 27.8\% of pairs and Gemini3.5-Flash in 88.9\%.

\textbf{Input Modality Ablation.}
Fig.~\ref{fig:modality_ablation} compares two input changes with the native audio-video input.
\textit{+ Text Query} appends the annotated target utterance as text.
\textit{Audio $\rightarrow$ ASR} replaces the audio with a timestamped transcript of all speech.
Of the 30 configurations, 27 improve with \textit{+ Text Query} and 21 with \textit{Audio $\rightarrow$ ASR}.

\textit{+ Text Query} spares a model from identifying the request in speech.
The weakest selected models gain the most, suggesting that request identification is their main bottleneck.
The strongest models change little, as they already identify most requests from speech.

\textit{Audio $\rightarrow$ ASR} makes the words explicit but drops acoustic cues such as emotion, intonation, speaker identity, and background sound.
The weakest models improve, suggesting that they poorly understand speech in raw audio.
The mid-ranked models decline, suggesting that the lost acoustic cues outweigh the benefit of explicit words.
The strongest models keep or improve their scores.
All of them run in thinking mode, which may help recover the request from the transcript and the video.

%% file: figure/fig_category_radar.tex
\begin{figure}[t]
    \centering
    \includegraphics[width=\linewidth]{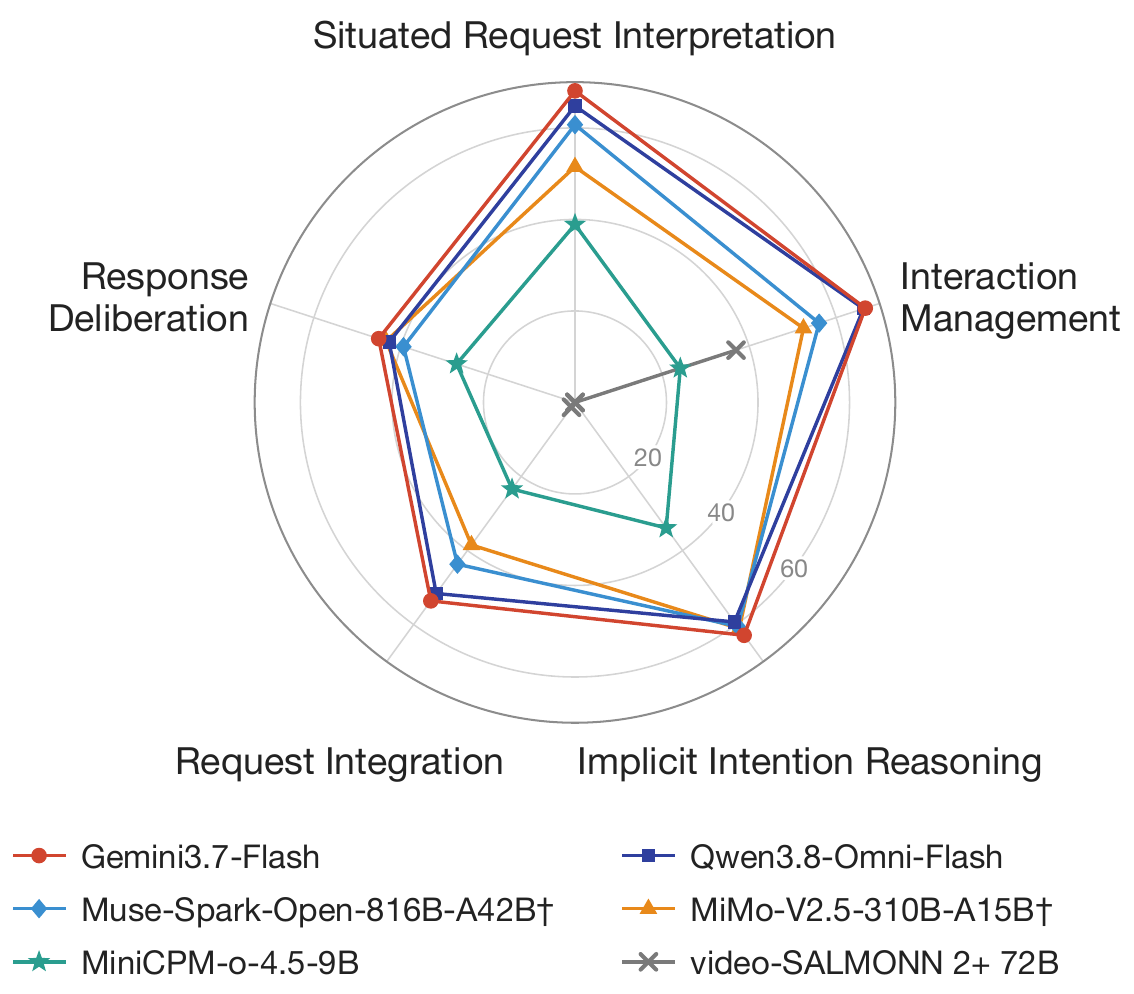}
    \caption{\textbf{Scores across five task categories.} Category scores for selected models are weighted by subcategory sample counts.}
    \label{fig:category_radar}
\end{figure}

%% file: figure/fig_3level_histogram.tex
\begin{figure}[t]
    \centering
    \includegraphics[width=\linewidth]{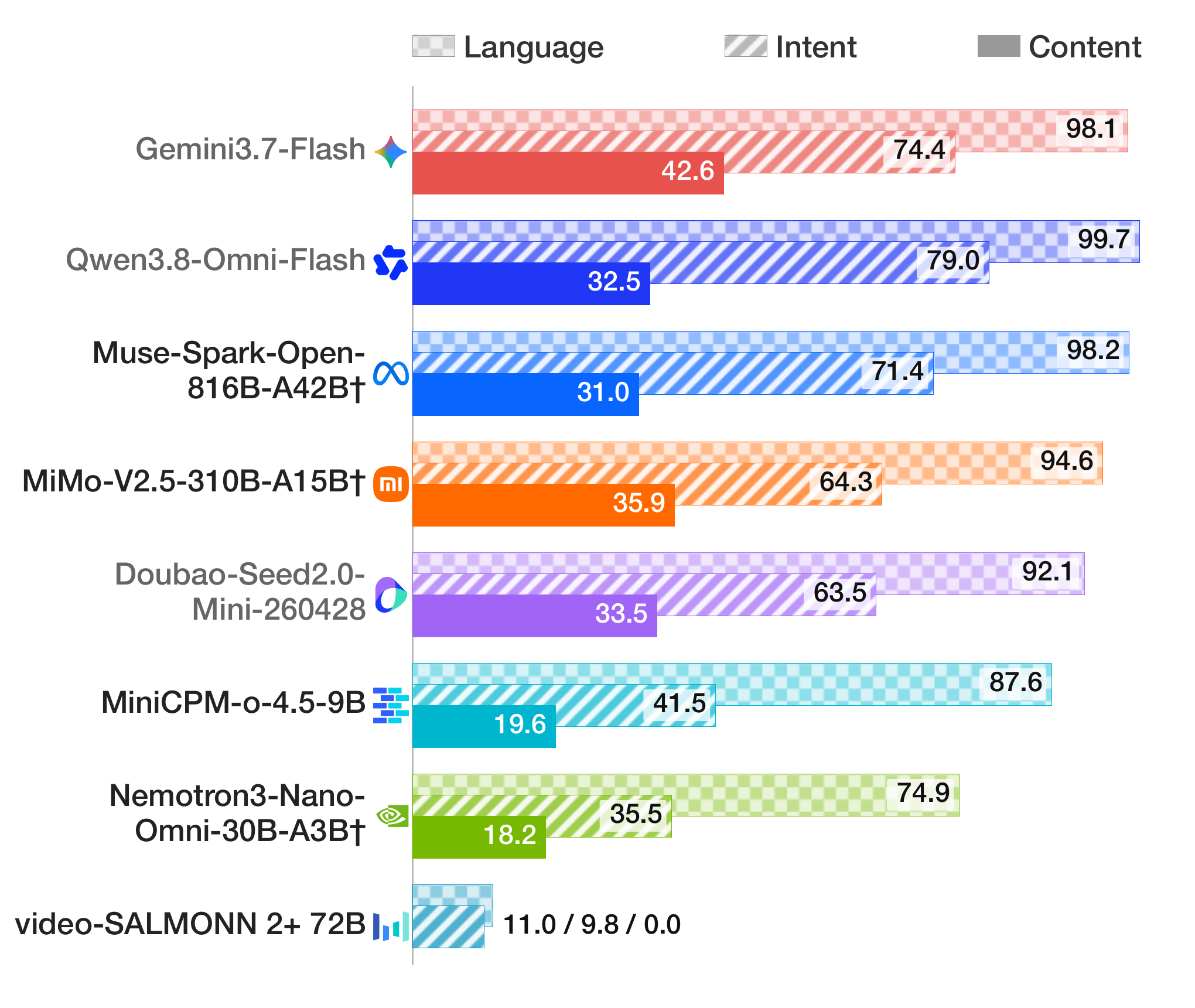}
    \caption{\textbf{Language, intent, and content pass rates of selected models.} Gray names denote proprietary models.}
    \label{fig:3level_histogram}
\end{figure}

%% file: figure/fig_overall_language_comparison.tex
\begin{figure}[t]
    \centering
    \includegraphics[width=\linewidth]{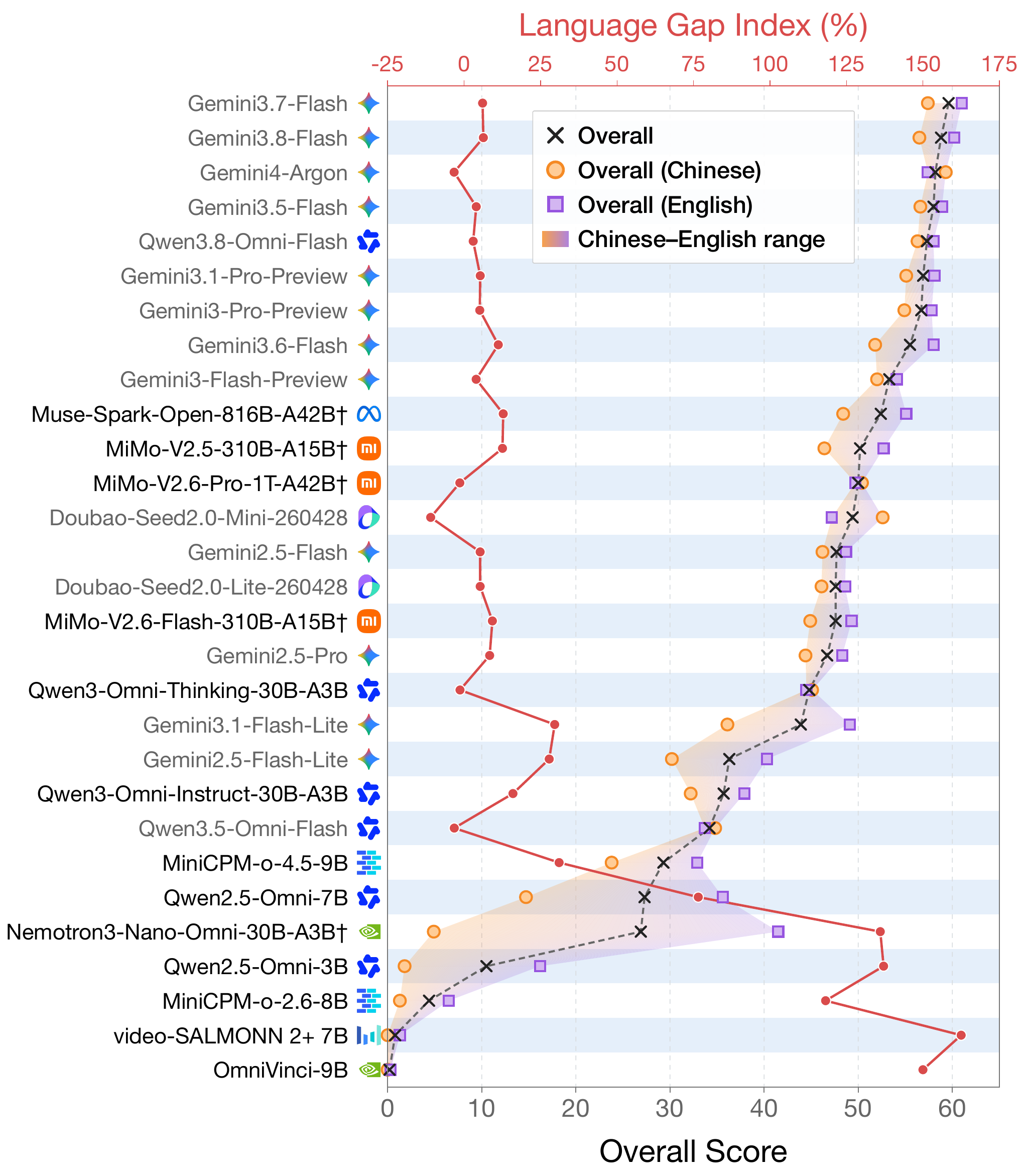}
    \caption{\textbf{Overall scores by language.}}
    \label{fig:overall_language_comparison}
\end{figure}

%% file: figure/fig_silence_overall.tex
\begin{figure}[t]
    \centering
    \includegraphics[width=\linewidth]{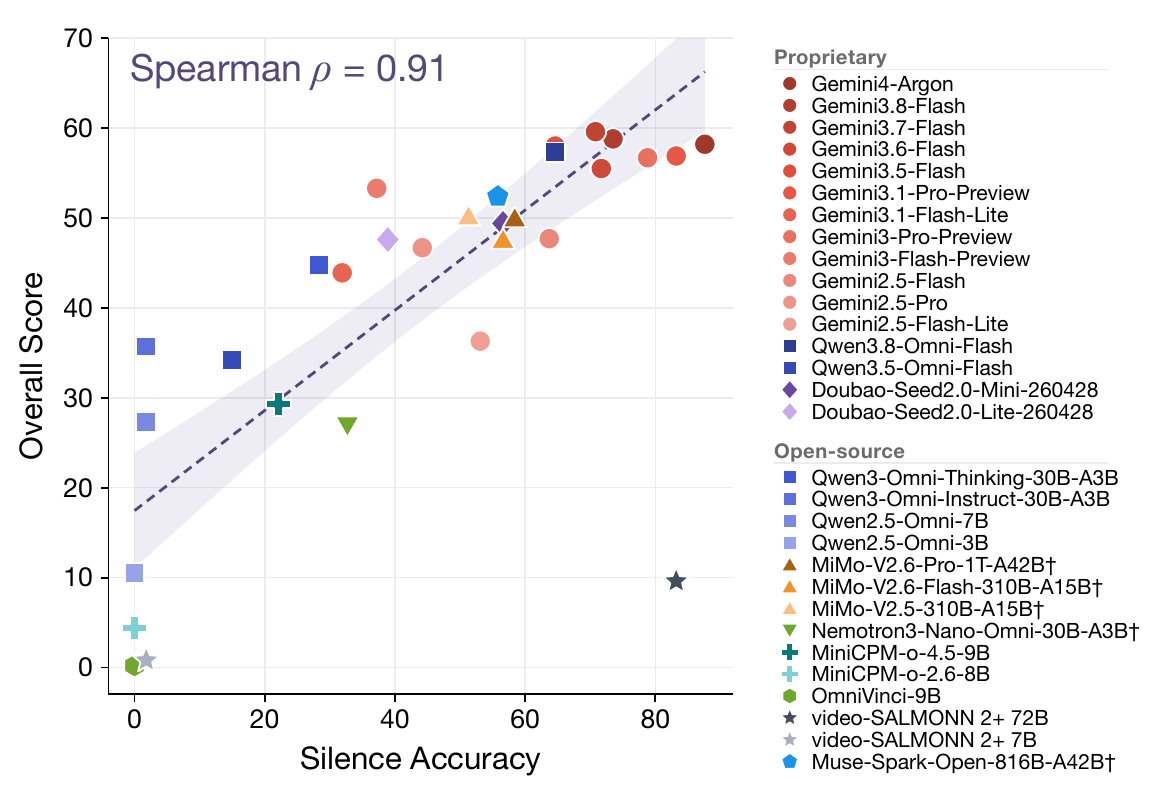}
    \caption{\textbf{Silence accuracy vs.\ Overall score}, with Spearman correlation of 0.91 (video-SALMONN 2+ 72B excluded).}
    \label{fig:silence_overall}
\end{figure}

%% file: figure/fig_modality_ablation.tex
\begin{figure*}[t]
    \centering
    \includegraphics[width=\linewidth]{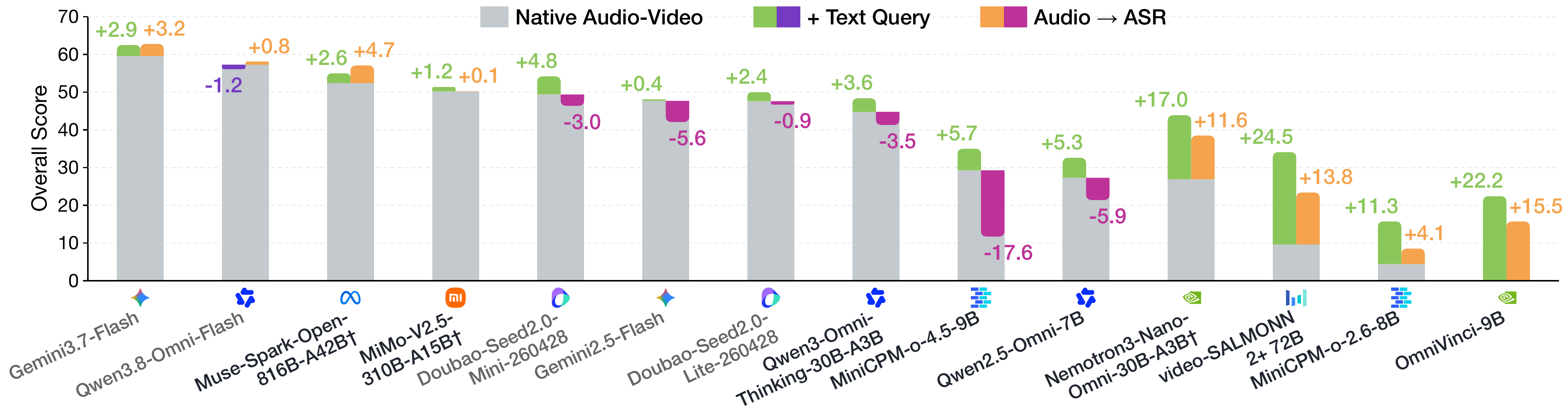}
    \caption{\textbf{Overall changes from native audio-video input} (gray bars) under \textit{+ Text Query} (left) and \textit{Audio $\rightarrow$ ASR} (right). Dark gray names denote proprietary models.}
    \label{fig:modality_ablation}
\end{figure*}

%% file: table/table_contrastive_pairs.tex
\begin{table}[t]
\centering
\caption{\textbf{Response behavior on contrastive pairs (\%).}
\textit{Exp.} is the expected \textit{Both} rate if the two samples were answered independently.
\textit{Same Decision} counts pairs given the same decision although one sample requires silence.}
\label{tab:contrastive_pairs}
\footnotesize
\setlength{\tabcolsep}{3pt}
\renewcommand{\arraystretch}{1.1}
\resizebox{\linewidth}{!}{%
\begin{tabular}{@{}l c c cc c rrrr@{}}
\toprule
\multirow{2}{*}{\textbf{Model}} & \multirow{2}{*}{\textbf{Overall}} & & \multicolumn{2}{c}{\textbf{Pair Accuracy}} & & \multicolumn{4}{c}{\textbf{Same Decision} $\downarrow$} \\
\cmidrule{4-5} \cmidrule{7-10}
& & & \textit{Both} & \textit{Exp.} & & \textit{MSA} & \textit{AST} & \textit{CA} & \textit{All} \\
\midrule
Gemini3.7-Flash    & 59.6 & & 20.4 & 20.5 & & 88.9 &  5.3 & 20.0 & 40.4 \\
Gemini3.8-Flash    & 58.8 & & 21.1 & 21.1 & & 66.7 & 10.5 & 30.0 & 36.2 \\
Gemini4-Argon      & 58.2 & & 23.1 & 22.7 & & 27.8 &  0.0 & 10.0 & 12.8 \\
Gemini3.5-Flash    & 58.0 & & 17.0 & 17.9 & & 88.9 & 10.5 & 30.0 & 44.7 \\
Qwen3.8-Omni-Flash & 57.3 & & 15.6 & 12.9 & & 83.3 & 15.8 & 30.0 & 44.7 \\
\midrule
Muse-Spark-Open\textsuperscript{\textdagger}     & 52.4 & & 7.5 & 10.6 & &  77.8 &  36.8 & 40.0 & 53.2 \\
MiMo-V2.5\textsuperscript{\textdagger}           & 50.2 & & 8.2 & 10.3 & &  66.7 &  42.1 & 60.0 & 55.3 \\
Qwen3-Omni-Thinking\textsuperscript{\textdagger} & 44.8 & & 5.4 &  5.7 & & 100.0 &  94.7 & 40.0 & 85.1 \\
Qwen3-Omni-Instruct                              & 35.7 & & 0.7 &  1.5 & & 100.0 & 100.0 & 90.0 & 97.9 \\
Qwen2.5-Omni-7B                                  & 27.3 & & 3.4 &  2.6 & & 100.0 &  94.7 & 100.0 & 97.9 \\
\midrule
\textit{Median of All models}                            & --   & & 7.1 &  8.9 & &  88.9 &  39.5 & 50.0 & 59.6 \\
\bottomrule
\end{tabular}%
}
\end{table}

%% file: sec/6_conclusion.tex
We have introduced \ourbench to evaluate whether omni-modal LLMs understand spoken intentions.
It keeps each request in native speech and the scene, and judges each response with a rubric verified against the video.
Our evaluation of 30 configurations suggests that the answer to the title question is not yet.
Current models do not always recognize complex requests, and they often keep responding when the speech is a conversation between peers rather than a request to the assistant.
Even after recognizing a request, they often fail to satisfy gestural constraints, the needs of several people, or the limits of the available evidence.
Recognizing a request and meeting its requirements are thus separate abilities, and an omni-modal assistant should treat the shared scene as part of the request.